\documentclass{article}

\usepackage[accepted]{mlsys2024}
\usepackage{amsmath,amsfonts}
\usepackage{graphicx}
\usepackage{textcomp}
\usepackage{xcolor}
\usepackage[utf8]{inputenc}
\usepackage{amsfonts}
\usepackage{enumitem}
\usepackage{caption}
\usepackage{subcaption}
\usepackage[export]{adjustbox}
\usepackage{bbding}
\usepackage{makecell}
\usepackage{float}

\usepackage{amsmath}

\usepackage{clrs+,listings, url}
\usepackage{hyperref,xcolor}

\usepackage{graphicx, color}

\newcommand{\revision}[1]{\textcolor{black}{#1}}

\newcommand{\mA}{\mathbf{A}}

\newcommand{\mW}{\mathbf{W}}

\newcommand{\mH}{\mathbf{H}}

\newcommand{\mQ}{\mathbf{Q}}

\newcommand{\vC}{\mathbf{c}}

\newcommand{\mP}{\mathbf{P}}

\begin{document}

\twocolumn[
\mlsystitle{Distributed Matrix-Based Sampling for Graph Neural Network Training}

\begin{mlsysauthorlist}
\mlsysauthor{Alok Tripathy}{ucb,lbl}
\mlsysauthor{Katherine Yelick}{ucb,lbl}
\mlsysauthor{Ayd{\i}n Bulu\c{c}}{ucb,lbl}
\end{mlsysauthorlist}
\mlsysaffiliation{ucb}{University of California, Berkeley}
\mlsysaffiliation{lbl}{Lawrence Berkeley National Laboratory}
\mlsyscorrespondingauthor{Alok Tripathy}{alokt@berkeley.edu}

\mlsyskeywords{graph neural networks, high-performance computing}
\vskip 0.3in
\begin{abstract}%

Graph Neural Networks (GNNs) offer a compact and computationally efficient way to learn embeddings and classifications on graph data. GNN models are frequently large, making distributed minibatch training necessary.

The primary contribution of this paper is new methods for reducing communication in the sampling step for distributed GNN training. Here, we propose a \textit{matrix-based bulk sampling} approach that expresses sampling as a sparse matrix multiplication (SpGEMM) and samples multiple minibatches at once. 
When the input graph topology does not fit on a single device, our method distributes the graph and use communication-avoiding SpGEMM algorithms to scale GNN minibatch sampling, enabling GNN training on much larger graphs than those that can fit into a single device memory. When the input graph topology (but not the embeddings) fits in the memory of one GPU, our approach (1) performs sampling without communication, (2) amortizes the overheads of sampling a minibatch, and (3) can represent multiple sampling algorithms by simply using different matrix constructions. In addition to new methods for sampling, we introduce a pipeline that uses our matrix-based bulk sampling approach to provide end-to-end training results. We provide experimental results on the largest Open Graph Benchmark (OGB) datasets on $128$ GPUs, and show that our pipeline is $2.5\times$ faster than Quiver (a distributed extension to PyTorch-Geometric) on a $3$-layer GraphSAGE network. On datasets outside of OGB, we show a $8.46\times$ speedup on $128$ GPUs in per-epoch time. Finally, we show scaling when the graph is distributed across GPUs and scaling for both node-wise and layer-wise sampling algorithms. 
\end{abstract}]
%

\printAffiliationsAndNotice{}

\section{Introduction}
Graph Neural Networks (GNNs) are a class of neural networks increasingly used for scientific and industrial problems including protein family classification in proteomics, track reconstruction in particle tracking, and fraud detection~\cite{wu2020comprehensive}

Training GNNs requires sampling minibatches from the training set. Minibatch training is more common than full-batch because it yields faster time-to-convergence and higher accuracy. For node classification, however, sampling batches is more complex than other deep learning (DL) problems since vertices in a batch have edges to vertices outside the batch. Training each batch of vertices for an $L$-layer GNN accesses the entire $L$-hop neighborhood of the batch (\textit{neighborhood explosion}), which is prohibitively expensive for many real-world graphs. 

Many have proposed algorithms that sample from the $L$-hop neighborhood of a minibatch. These algorithms are necessary, and effectively reduce the time and memory costs to train a single minibatch. Broadly, sampling algorithms can be categorized as either: (1) node-wise, (2) layer-wise, and (3) graph-wise. 

Unfortunately, sampling algorithms must be able to access the entire graph per batch, which is too large to fit in GPU memory. Current GNN tools, such as Deep Graph Library (DGL)~\cite{zheng2020distdgl} and PyTorch Geometric (PyG)~\cite{pyg}, run sampling on CPU as a consequence — a weaker processor compared to GPUs. Those that do not, such as Quiver, restrict the graph size by fully replicating the graph on each GPU used. In addition, sampling algorithms are frequently the bottleneck of GNN training when compared to forward and backward propgation, and only GraphSAGE has an existing multi-node implementation. In this work, we propose a method to sample on a graph distributed across GPUs, and show its effectiveness across many types of sampling algorithms.

Distributed and GPU graph analytics are well-studied fields, and expressing graph algorithms in the language of sparse linear algebra is a proven approach~\cite{bulucc2011combinatorial,yang2022graphblast}. By representing a graph algorithm in terms of matrix operations, one can leverage existing decades of work in distributed sparse matrix algorithms to distribute graph computation. 

In this work, we show how to express node-wise and layer-wise sampling algorithms in the language of linear algebra. We propose a matrix-based bulk sampling approach that (1) amortizes the cost of sampling a minibatch, and (2) leverages distributed, communication-avoiding sparse matrix multiplication algorithms for scalability. We wrap our sampling step in an end-to-end training pipeline, and present this pipeline's performance results. 
As a byproduct, we also introduce the first fully distributed implementation of the LADIES algorithm. We provide theoretical and empirical results that outperform existing GNN tools.

\section{Background}
\begin{table}[H]
\centering
\footnotesize
\caption{List of symbols and notations used in our pipeline} \label{symboltable}
\vspace{-0.2cm}
\begin{tabular}{ |p{2cm}||p{5.5cm}|}
\hline
\multicolumn{2}{|c|}{Symbols and Notations} \\
\hline
Symbol & Description  \\
\hline
$\mA$ & Adjacency matrix of graph ($n \times n$)\\
$\mH^l$ & Embedding matrix in layer $l$ ($n \times f$)\\
$\mW^l$ & Weight matrix in layer $l$ ($f \times f$)\\
$\mQ^l$ & Sparse sampler matrix in layer $l$ \\
$\mP$ & Probability matrix during sampling \\
$\mA_{S}$ & Sampled adjacency matrix \\
$f$ & Length of feature vector per vertex \\
$L$ & Total layers in GNN \\
$p$ & Total number of processes \\
$\alpha$ & Latency \\
$\beta$ & Reciprocal bandwidth \\
$b$ & Batch size \\
$s$ & Sampling parameter \\
$k$ & Number of batches to sample in bulk \\
\hline 
\end{tabular}
\vspace{-0.3cm}
\end{table}

\subsection{Graph Neural Networks}
Graph Neural Networks take as input a graph $G = (V, E)$. While GNNs can solve a wide variety of machine learning problems, we focus on \textit{node classification} without loss of generality. In this problem, each vertex takes an associated \textit{feature vector} as input, and a subset of vertices have an associated \textit{label}. The objective of the network is to classify unlabelled vertices in the graph using input features, graph connectivity, and vertex labels. 

GNNs follow the \textit{message-passing} model, consisting of a \texttt{message} step and an \texttt{aggregate} step per iteration of training~\cite{hamiltonInductive2017}. The \texttt{message} step creates a message per edge in the graph. The \texttt{aggregate} step takes a vertex $v$ and combines the messages across all of $v$'s incoming neighbors. The output is multiplied with a parameter weight matrix, and the result is an embedding vector $z_v$. After several layers of these steps, the network outputs an embedding vector per vertex, after which the network inputs vectors and labels into a loss function for backpropagation. Formally, for an arbitrary layer $l$, computing an embedding vector $z^l_v$ is
$$z^l_v = \text{AGG}(\text{MSG}(z^{l-1}_u, z^{l-1}_v)) \forall u\in N(v), 
z^l_v = z^l_v\mW^l$$


While the size of the batch is small compared to the vertex set, running message-passing on a batch naively touches a large fraction of the input graph. Training a batch $B$ in an $L$-layer network requires accessing the $L$-hop aggregated neighborhood of $B$. This phenomenon is referred to as \textit{neighborhood explosion}. To alleviate costs, minibatch training GNNs includes a \textit{sampling step} that samples the $L$-hop neighborhood of each batch. Message-passing on a sampled batch will only aggregate from vertices in its sampled $L$-hop neighborhood. The specific sampling algorithm depends on the problem, and there are many such algorithms in the literature. In addition, in our notation, the $L$-th layer contains the batch vertices, and the 1st layer has vertices furthest from the batch vertices.

\begin{figure}[!t]
    \centering
    \includegraphics[scale=0.3]{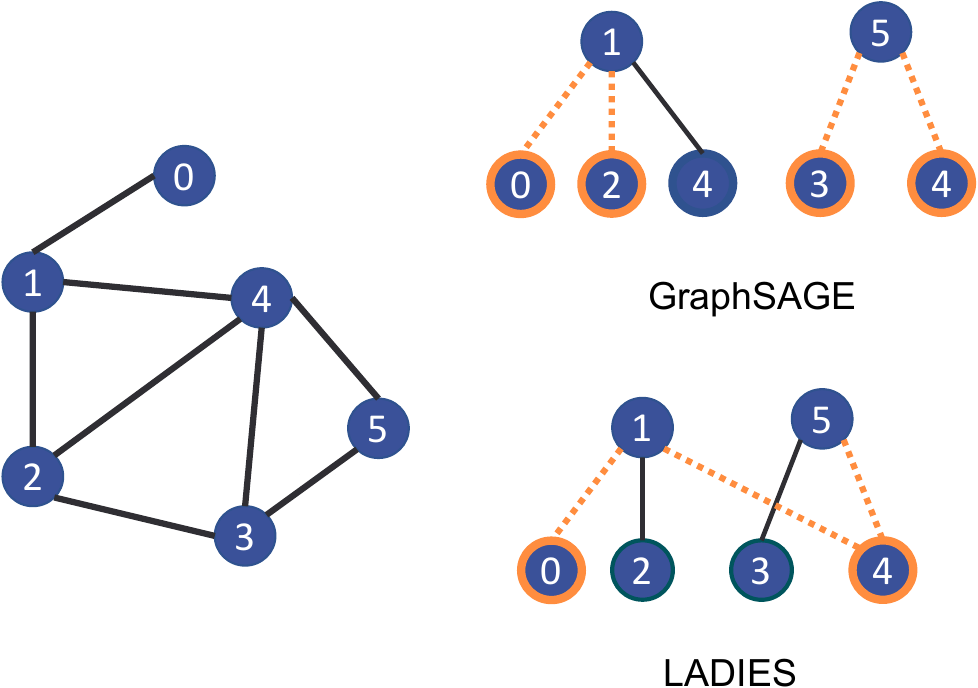}
    \caption{Example outputs for both GraphSAGE and LADIES sampling on a batch with vertices $\{1, 5\}$ and a sample number of $s=2$. Bolded vertices denote vertices included in the sample, and dashed edges denote edges included in the sample.}
    \label{fig:examples}
    \vspace{-0.5cm}
\end{figure}
\subsection{Sampling Algorithms}
GNN sampling algorithms can broadly be classified into three taxonomies with various tradeoffs: 1) node-wise sampling, 2) layer-wise sampling, and 3) graph-wise sampling. We focus on node-wise and layer-wise sampling in this work, hence we will only give background on those taxonomies. 
\subsubsection{Node-wise sampling}
Node-wise algorithms sample from an $L$-hop neighorhood by sampling vertices in the $L$-hop neighborhood of each individual batch vertex. 

{\bf GraphSAGE:} GraphSAGE is the simplest and most widely used example of node-wise sampling~\cite{hamiltonInductive2017}. In GraphSAGE, each vertex in the batch samples $s$ of its neighbors as the next layer in the batch uniformly at random. The sample number, $s$, is a hyperparameter. For a multi-layer GNN, each vertex in a layer will sample $s$ of its neighbors for the next layer as well. Figure~\ref{fig:examples} illustrates this for the example graph and $s=2$. GraphSAGE is a simple algorithm that is straightforward to implement. In our work, we show how our linear algebraic sampling framework expresses node-wise sampling by implementing GraphSAGE. 
\subsubsection{Layer-wise sampling}
Layer-wise algorithms sample from an $L$-hop neighborhood by sampling a set of vertices $S^l\subset V$ per layer, and including every edge between $S^{l+1}$ and $S$. The simplest layer-wise sampling algorithm is FastGCN~\cite{chen2018fastgcn}. FastGCN samples $s$ vertices from $V$, where each vertex is sampled with a probability correlated with its degree in $G$. Since the number of vertices in a layer is constrained to $s$, FastGCN avoids  neighborhood explosion. However, note that vertices in $S$ are not necessarily in the aggregated neighborhood of $S^{l+1}$, which affects accuracy when training with FastGCN. 

{\bf LADIES:} Zou et. al. introduce Layer-Wise Dependency Sampling (LADIES) to ensure only vertices in the aggregated neighborhood of the previous layer are sampled~\cite{ladies}. The algorithm sets the probability a vertex is selected to be correlated with the number of neighbors it has in $S^l$. If $e_v$ is the number of neighbors vertex $v$ has in $S$, then the probability $v$ is sampled is
$p_v = \frac{e_v^2}{\sum_{u\in V}e_u^2}$.
In Figure~\ref{fig:examples}, for a batch $\{1, 5\}$, the probability array for all $6$ vertices is
$
\Big[\frac{1}{7}, 0, \frac{1}{7}, \frac{1}{7}, \frac{4}{7}, 0\Big]
$
In the example, we have $s=2$ and sampled vertices $\{0, 4\}$. Consequently, the sample for LADIES includes every edge between $\{1, 5\}$ and $\{0, 4\}$. In our work, we show how our linear algebraic sampling framework expresses layer-wise sampling by implementing LADIES.

\subsection{Distribution Sampling}
All GNN sampling algorithms require sampling $s$ elements from some probability distribution. Two common distributed sampling algorithms are \textit{inverse transform sampling (ITS)} and \textit{rejection sampling}~\cite{pandey2020csaw,knightking,olver2013fast,schneider2022parallel}. Rejection sampling risks taking many iterations to complete, while ITS uses a prefix sum. In our work, we use ITS, and empirically show the prefix sum is a negligible cost in our problem. 
\subsection{Notation}
Table~\ref{symboltable} lists the notation we use. Layer $L$ refers to the last layer in the network with only the vertices in the minibatch. In addition, we use the $\alpha-\beta$ model to analyze communication costs. Here, each message takes a constant $\alpha$ time units latency irrespective of its size plus an inverse bandwidth term that takes $\beta$ time units per word in the message. Thus, sending a message of $k$ words takes $\alpha + \beta k$ time. 
\section{Related Work}
\subsection{Distributed GNN Systems}
Real-world graphs and GNN datasets are frequently too large to fit on a single device, necessitating distributed GNN training. The two most popular GNN training tools are Deep Graph Library (DGL) and PyTorch Geometric (PyG), which can be distributed with Quiver~\cite{quiver}. Both DGL and Quiver run most sampling algorithms on CPU with the graph stored in RAM. For node-wise sampling, these tools support sampling with the graph stored on GPU.

In addition to these, there exist many distributed GNN systems for both full-batch and minibatch training. Full-batch training systems include NeuGraph~\cite{neugraph}, ROC~\cite{mlsys2020_83}, AliGraph~\cite{zhu2019aligraph}, CAGNET~\cite{tripathy2020reducing}, PipeGCN~\cite{wan2022pipegcn}, BNS-GCN~\cite{wan2022bns}, and CoFree-GNN~\cite{cao2023communicationfree}. In full-batch distributed GNN training, the graph is typically partitioned across devices, and the main performance bottleneck is communicating vertex embeddings between devices. Each work varies in their approaches to reduce this communication. 

While full-batch training smoothly approaches minima in the loss landscape, minibatch training tends to achieve better generalization by introducing noise. Thus, in this work, we focus on minibatch training.
For minibatch training, existing systems include DistDGL~\cite{zheng2020distdgl}, Quiver, GNNLab~\cite{yang2022gnnlab}, WholeGraph~\cite{yang2022wholegraph}, DSP~\cite{cai2023dsp}, PGLBox~\cite{pglbox}, SALIENT++~\cite{kaler2023salient++}, NextDoor~\cite{jangda2021accelerating}, $P^3$~\cite{gandhi2021p3}. Here, the main performance bottleneck is the cost of sampling minibatches due to random memory accesses, difficulty parallelizing, and communicating samples from CPU to GPU, with studies showing that sampling can take up to 60\% of the total training time~\cite{jangda2021accelerating,yang2022gnnlab}. 

Notably, none of the minibatch systems in the current literature achieve all of the following: (1) sample minibatches on GPU, (2) support distributed, multi-node training, and (3) supports multiple sampling algorithms (Table ~\ref{tab:rwtable}). Sampling on GPUs, if possible, is preferred to CPUs as it avoids communicating samples from CPU to GPU over a low-bandwidth link like PCIe. In addition, multi-node training is necessary to train large GNN datasets. Some systems, such as Quiver and WholeGraph, support multi-node training by replicating the graph dataset (topology and embeddings) on each node. While this replication can take advantage of additional compute resources, it still limits the size of the dataset for training. Finally, while most systems support node-wise sampling algorithms, layer-wise and graph-wise sampling algorithms have shown to achieve higher accuracy for certain applications. Supporting multiple types of sampling algorithms on GPU is necessary for a robust GNN training system. 

\subsection{Graph Sampling Systems}
In addition to GNN systems, many have worked on systems that return samples of an input graph. KnightKing is a distributed CPU-based graph sampler, and C-SAW is a GPU-based graph sampler~\cite{knightking,pandey2020csaw}. These systems address the broader problem of graph sampling, while our work is tailored towards sampling algorithms in the context of GNN training.

\begin{table}[t]
\begin{small}
\begin{center}
\footnotesize
\caption{Existing distributed minibatch GNN systems} 
\vspace{-0.2cm}
\begin{tabular}{ |p{2.5cm}||c|c|c|}
\hline
\multicolumn{4}{|c|}{Distributed Minibatch GNN Systems} \\
\hline
\thead{System} & \thead{GPU\\ Sampling} & \thead{Multi-node \\ Training*} & \thead{Multiple\\ Samplers}\\
\hline
DistDGL & \CheckmarkBold & \CheckmarkBold & \\
\hline
Quiver & \CheckmarkBold & \CheckmarkBold & \\
\hline
GNNLab & \CheckmarkBold & & \\
\hline
WholeGraph & \CheckmarkBold & & \\
\hline
DSP & \CheckmarkBold & & \CheckmarkBold \\
\hline
PGLBox & \CheckmarkBold & & \\
\hline
SALIENT++ & & \CheckmarkBold & \\
\hline
NextDoor & \CheckmarkBold & & \CheckmarkBold \\
\hline
$P^3$ & & \CheckmarkBold & \\
\hline
\textbf{This work} & \CheckmarkBold & \CheckmarkBold & \CheckmarkBold \\
\hline
\end{tabular}
\label{tab:rwtable}
\end{center}
\footnotesize{$*$ does not include systems that require replicating both the graph and features on each node or GPU, as this limits the datasets that can be trained.}\\
\end{small}
\vspace{-0.8cm}
\end{table}

\begin{figure*}[h]
    \centering
    \begin{subfigure}[t]{0.42\textwidth}
        \includegraphics[width=\textwidth]{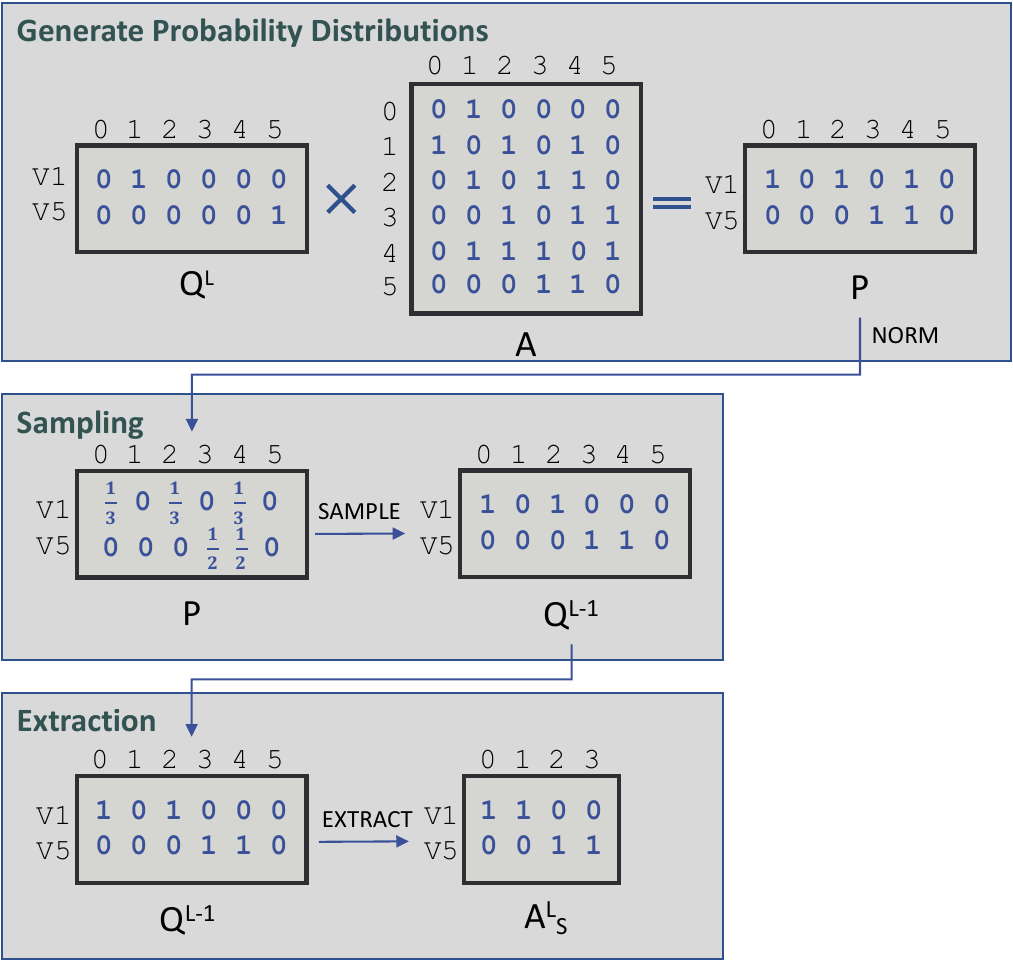}
        \caption{GraphSAGE Operations}
        \label{fig:graphsage_figure}
    \end{subfigure}
    \hfill
    \begin{subfigure}[t]{0.42\textwidth}
        \includegraphics[width=\textwidth]{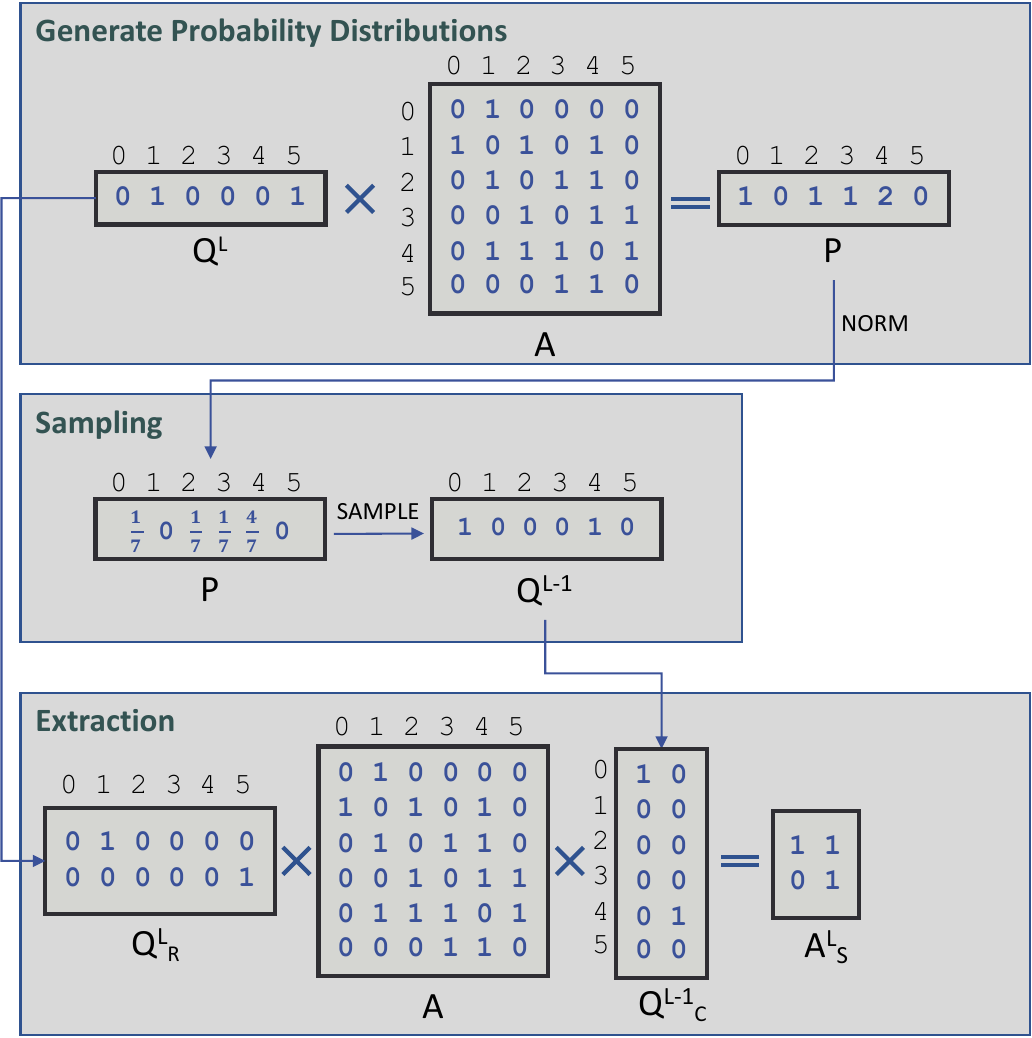}
        \caption{LADIES Operations}
        \label{fig:ladies_figure}
    \end{subfigure}
    \caption{Diagram of matrix operations used to sample the first layer of a GNN formed from the example graph in Figure~\ref{fig:examples} along with the minibatch $\{1, 5\}$.}
    \vspace{-0.5cm}
\end{figure*}

\section{Matrix-Based Algorithms for Sampling}
\label{sec:matrixbased}
In this section, we introduce our approach to represent GraphSAGE and LADIES with matrix operations, and how we use this formulation to sample minibatches in bulk. We first discuss how to sample a single minibatch with our matrix-based approach. Subsequently, we show how to generalize to sampling multiple minibatches in bulk. We discuss how to distribute these algorithms in Section~\ref{sec:distributed}, and only focus on the matrix representations in this section. 

Each algorithm takes as input
\setlist{nolistsep}
\begin{enumerate}[noitemsep]
    \item $\mA \in \{0,1\}^{n \times n}:$ sparse adjacency matrix,
    \item $\mQ^{L}:$ sparse sampler-dependent matrix,
    \item $b:$ batch size,
    \item $s:$ sampling parameter,
\end{enumerate}
The output of sampling is a list of sampled adjacency matrices $\mA^0\ldots\mA^{L-1}$ for a minibatch. Each sampled adjacency matrix $\mA^{l}$ is the adjacency matrix used in layer $l$'s aggregation step in forward propagation. $\mQ^L$ is a sparse matrix that holds the minibatch vertices. The exact structure and dimensions for $\mQ^L$ is dependent on the sampling algorithm, and will be specified in the respective subsection.

\begin{algorithm}[t]

\caption{Matrix-based abstraction for sampling algorithms. Each algorithm implements the NORM and EXTRACT functions. SAMPLE uses Inverse Transform Sampling.}
\label{alg:framework}
\begin{algorithmic}[1]
\FOR{$l = L$ {\bfseries to} $1$}
    \STATE $\mP \gets \mQ^l\mA$ \label{line:genprobdist1}
    \STATE $\mP \gets$ NORM($\mP$) \label{line:genprobdist2}
    \STATE $\mQ^{l-1} \gets$ SAMPLE($\mP, b, s$) \label{line:sample}
    \STATE $\mA^{l} \gets$ EXTRACT($\mA, \mQ^l, \mQ^{l-1}$) \label{line:extract}
\ENDFOR
\end{algorithmic}
\end{algorithm}

Each algorithm follows the same three-step framework outlined in Figure~\ref{fig:graphsage_figure} --- 1) generate probability distributions, 2) sample from each distribution, 3) extract rows and columns from the input adjacency matrix. $\mP$ holds one distribution per row, and the sampling step reduces to sampling from each row of $\mP$. For GraphSAGE, each row of $\mP$ is the neighborhood of a batch vertex. For LADIES, each row of $\mP$ is the distribution across the aggregated neighborhood of a batch. Once we compute $\mP$, we sample $s$ nonzeros per row of $\mP$ using ITS. We then extract rows of $\mA$ from the vertices sampled in the prior layer ($\mQ^l$) and columns of $\mA$ from the newly sampled vertices ($\mQ^{l-1}$). In the respective GraphSAGE and LADIES subsections, we detail the structure for $\mQ^L$, and the matrix operations used to implement each function call in Algorithm~\ref{alg:framework}.

\subsection{GraphSAGE}
\subsubsection{Generating Probability Distributions}
In GraphSAGE, each batch vertex samples $s$ of its neighbors. Thus, for a single batch, $\mP$ has $b$ rows, one for each batch vertex $v_0,\ldots,v_{b-1}$ and $n$ columns in the first iteration of Algorithm~\ref{alg:framework}. Each row $i$ has a nonzero for each respective neighbor of vertex $v_i$. In addition, each nonzero value in row $i$ is $1/|N(v_i)|$.

To compute $\mP$, our input matrix $\mQ^L$ has the same dimensions as $\mP$. $\mQ^L$ has a single nonzero per row, where the nonzero for row $i$ is in column $v_i$. We then multiply $\mP \gets \mQ^L\mA$ as an SpGEMM operation, and divide each nonzero of $\mP$ by the sum of its row. Figure~\ref{fig:graphsage_figure} depicts this for the example graphs in Figure~\ref{fig:examples}. In the first layer of sampling, the dimensions of $\mP$ and $\mQ^L$ are both $b\times n$. As sampling continues, the dimensions become $bs^l\times n$ for each layer $l$, for each matrix.
%

%

\subsubsection{Sampling from Distributions}
\revision{We use ITS to sample from each distribution. Our sampling code takes matrix $\mP$ as input, which has one distribution per row, and outputs a matrix $\mQ^l$ with exactly $s$ nonzeros per row. To construct $\mQ^l$ with ITS, we first run a prefix sum per row of $\mP$. We then generate $s$ random numbers per row of $\mP$, and binary search each number within its row's prefix sum to select $s$ separate nonzero columns (i.e. sampled vertices). This process repeats to select $s$ distinct nonzero elements per row of $\mP$ without replacement, forming $\mQ^l$ as our output matrix. Figure~\ref{fig:graphsage_figure} shows the output $\mQ^l$ matrix after ITS sampling the example $\mP$ matrix.} 

\subsubsection{Row and Column Extraction}
To construct the final sampled adjacency matrix $\mA^l$ for a layer $l$, we only need to remove empty columns in $\mQ^{l-1}$. Each sampled matrix in $\mA_S$ stores the edges connecting batch vertices to sampled vertices in the next layer. Since each batch vertex is a row of $\mQ^{l-1}$, the edges connecting each batch vertex already exist in its row. Figure~\ref{fig:graphsage_figure} shows the example $\mQ^{L-1}$ matrix getting extracted to form its sampled adjacency matrix. Note that the number of rows of $\mA^l$ is $bs^l$, and the number of columns is $bs^{l+1}$.
\subsubsection{Bulk GraphSAGE Sampling}
To sample a set of minibatches with our matrix approach, we can vertically stack the individual $\mQ^l, \mP$, and $\mA^l$ matrices across all minibatches. If $\mQ^l_i, \mP_i$, and $\mA^l_i$ are the matrices for a single batch $i$, and we have $k$ total batches, then to sample all $k$ batches we use Equation~\ref{eqn:pqstacking}.

The matrix operations as described above and in Algorithm~\ref{alg:framework} are identical for these stacked matrices as well. The dimensions for stacked $\mQ^l$ and $\mP^l$ are $kbs^l\times n$, while the dimensions for stacked $\mA^l$ are $kbs^l\times s^{l+1}$.
\begin{equation}
\mQ^l = \left( 
\begin{array}{c}
\mQ^l_{1} \\
\vdots \\
\mQ^l_{k}
\end{array} 
\right)
\mP = \left( 
\begin{array}{c}
\mP_{1} \\
\vdots \\
\mP_{k}
\end{array} 
\right)
\mA^l = \left( 
\begin{array}{c}
\mA^l_{1} \\
\vdots \\
\mA^l_{k}
\end{array} 
\right)
\label{eqn:pqstacking}
\end{equation}
\subsection{LADIES}
\label{sec:ladies}
\subsubsection{Generating Probability Distributions}
In LADIES, each batch samples $s$ vertices in the aggregated neighborhood of the batch. Thus, for a single batch, $\mP$ has one row and $n$ columns, with a nonzero in a column for each vertex in the aggregated neighborhood. Each nonzero value in a column $i$ is the probability vertex $i$ is selected, according the distribution defined by LADIES.

To compute $\mP$, our input matrix $\mQ^L$ has the same dimensions as $\mP$. $\mQ^L$ has a $b$ nonzeros in the row, each in a column per respective batch vertex. We then multiply $\mP \gets \mQ^L\mA$, and divide each nonzero of $\mP$ by the sum of its row. Figure~\ref{fig:ladies_figure} shows the input $\mQ^L$ used to compute the probability distribution for the example in Figure~\ref{fig:examples}.

\subsubsection{Sampling from Distributions}
We again use ITS to sample from each distribution. Like GraphSAGE, a row of $\mP$ in LADIES is a probability distribution from which we sample $s$ elements. Figure~\ref{fig:ladies_figure} shows the output of sampling the example in Figure~\ref{fig:examples}.

\subsubsection{Row and Column Extraction}
For LADIES, the sampled adjacency matrix for a single batch contains every edge connecting the set of batch vertices to the set of sampled vertices. We construct the sampled adjacency matrix by extracting the batch vertices' rows from $\mA$ and the sampled vertices' columns from $\mA$. Both steps can be expressed with row extract and column extraction SpGEMM operations.

We extract the batch vertex's rows from $\mA$ by converting $\mQ^L$ into a row extraction matrix $\mQ^L_R \in \{0,1\}^{b\times n}$, and multiplying $\mA_R \gets \mQ^L_R\mA$. $\mQ^L_R$ expands $\mQ^L$ to have each nonzero in one of $b$ rows, while keeping each nonzero's column id. Figure~\ref{fig:ladies_figure} shows the row extraction matrix $\mQ^L_R$ for the example graph in Figure~\ref{fig:examples}.

To run column extraction on a single batch, we construct a column  extraction matrix $\mQ^{l-1}_C \in \{0,1\}^{n\times s}$ and multiply $\mA_R\mQ^{l-1}_C$. $\mQ^{l-1}_C$ has one nonzero per column, at the row index of each vertex to extract from $\mA_R$. Figure~\ref{fig:ladies_figure} shows the column extraction $\mQ^L_C$ matrix for the example in Figure~\ref{fig:examples}. Multiplying $\mA_S \gets \mQ^L_R\mA\mQ^L_C$ yields our final sampled adjacency matrix for LADIES.

\subsubsection{Bulk LADIES Sampling}
To run LADIES and sample a set of minibatches with our matrix approach, we first vertically stack the individual $\mQ^l$, $\mP$, and $\mA^l$ matrices (similar to bulk sampling for GraphSAGE). If $\mQ^l_i, \mP_i$, and $\mA^l_i$ are the matrices for a single batch $i$, and we have $k$ total batches, then we can stack these matrices and have the same construction as Equation~\ref{eqn:pqstacking}.
The matrix operations for generating probability distributions and sampling from distributions, as described in the earlier parts of Section~\ref{sec:ladies}, are identical for these stacked matrices as well. 

In the bulk extraction step for LADIES, we cannot only stack our $\mQ^l_R$ and $\mQ^l_C$ matrices since extracting rows and columns must be separate for each minibatch. For row extraction, we stack the the $\mQ^l_R$ matrix across all minibatches and multiply this stacked matrix with $\mA$.
\begin{equation}
\mA_R = \left( 
\begin{array}{c}
\mA_{R1} \\
\vdots \\
\mA_{Rk}
\end{array}
\right)
=\left( 
\begin{array}{c}
\mQ^l_{R1} \\
\vdots \\
\mQ^l_{Rk}
\end{array} 
\right)
\mA \\
\label{eqn:rowstacking}
\nonumber
\end{equation}

For column extraction, we first expand $\mA_R$ into a block diagonal matrix, where each block on the diagonal is a separate $\mQ^l_{Ri}\mA$ product. 
$$
\mA_S = 
\left( 
\begin{array}{c c c}
\mA_{R1} & \ldots & 0 \\
\vdots & \ddots & \vdots \\
0 & \ldots & \mA_{Rk}
\end{array} 
\right)
\left( \\
\begin{array}{c}
\mQ^{l-1}_{C1} \\
\vdots \\
\mQ^{l-1}_{Ck}
\end{array} 
\right)
\label{eqn:colstacking}
$$
The final $\mA_S$ matrix is a stacked $kb\times s$ matrix, where each $b\times s$ submatrix is a sampled minibatch adjacency matrix.

\section{Distributed Sampling Algorithms}
\label{sec:distributed}
In this section, we introduce our distributed sparse matrix algorithms for both GraphSAGE and LADIES. We use two algorithms for distributed SpGEMM. Our first is a simpler algorithm that assumes that the input adjacency matrix $\mA$ can fit on device. Our second algorithm relaxes this constraint and partitions $\mA$ across devices, at the expense of extra communication. We use these algorithms for the $\mQ^l\mA$ SpGEMM when generating probability distributions, and for extraction in LADIES. For simplicity, we introduce these algorithms using $\mQ^l$ as the left matrix and $\mA$ as the right matrix, but these algorithms can be used when multiplying any two sparse matrices.
\subsection{Graph Replicated Algorithm}
In our first algorithm, we partition $\mQ^l$ across devices and replicate the adjacency matrix $\mA$ on each device. Our partitioning strategy for $\mQ^l$ is a \textit{1D block row distribution} on a process grid of size $p$. Here, $\mQ^l$ is split into $p$ block rows with each block row belonging to one process. Note that $\mQ^l$ could be a stacked matrix across $k$ minibatches, in which case each block row of $\mQ^l$ contains vertices in $k/p$ minibatches. If $\mQ^l_i$ is the block row of $\mQ^l$ belonging the process $P(i)$, the $\mQ^l\mA$ SpGEMM is equivalent to
\begin{equation}
\mQ^l\mA = \left( 
\begin{array}{c}
\mQ^l_{1} \\
\vdots \\
\mQ^l_{p}
\end{array} 
\right)
\mA
\label{eqn:1dpartitioning}
\nonumber
\end{equation}
Note that each process can compute its product $\mQ^l_i\mA$ locally without communication. This is true even if $\mQ^l$ is a stacked matrix. In addition, we eliminate communication in the sampling and extraction steps. For GraphSAGE, this is the case because both the sampling and extraction steps are row-wise operations, and each process stores a block row of $\mQ^l\mA$. For LADIES, the sampling is also row-wise, and the extraction step can fully replicate $\mA$ and $\mQ^l_C$ to ensure each SpGEMM's right matrix is replicated.
\subsection{Graph Partitioned Algorithm}
\label{sec:graphpartition}
When we partition the input graph $\mA$, we use a 1.5D partitioning scheme where $p$ processes are divided into a $p/c\times c$ process grid. Both input matrices $\mQ^l$ and $\mA$ are partitioned across this process grid~\cite{spdmmm16}. Here, $c$ is an input parameter known as the \textit{replication factor}. We decide to use a 1.5D scheme since prior work has shown 1.5D algorithms generally outperform other schemes (e.g. 1D, 2D, 3D) in other GNN and machine learning contexts~\cite{tripathy2020reducing,gholami2017integrated}.

With a 1.5D partitioning, both $\mQ^l$ and $\mA$ are partitioned into $p/c$ block rows, and each block row is replicated on $c$ processes. Specifically, each process in process row $P(i,:)$ stores $\mQ^l_i$ and $\mA_i$.
\begin{equation}
\mQ^l = \left( 
\begin{array}{c}
\mQ^l_{1} \\
\vdots \\
\mQ^l_{p/c}
\end{array} 
\right)
= \left(
\begin{array}{c c c}
\mQ^l_{11} & \ldots  & \mQ^l_{1, p/c} \\
\vdots  & \ddots  & \vdots  \\
\mQ^l_{p/c,1} & \ldots   & \mQ^l_{p/c, p/c} 
\end{array}, 
\right)
\nonumber
\end{equation}
\\
\begin{equation}
\mA = \left( 
\begin{array}{c}
\mA_{1} \\
\vdots \\
\mA_{p/c}
\end{array} 
\right)
\label{eqn:15dpartitioning}
\nonumber
\end{equation}
Given this partitioning for the inputs to our algorithm, we detail how to distribute each step of Algorithm~\ref{alg:framework} in its respective subsection. In addition, we provide a communication analysis for each step in GraphSAGE and LADIES. For space constraints, we restrict our analysis to just the SpGEMM when generating probability distributions. However, the analysis for extraction SpGEMMs follows the same process. 
\subsubsection{Generating Probability Distributions}
When generating probability distributions, the only step that requires communication is computing $\mP \gets \mQ^l\mA$. Normalizing $\mP$ is a row-wise operation, and $\mP$ is partitioned into block rows.
When computing $\mP \gets \mQ^l\mA$, each process row $P(i,:)$ computes $\mP_i = \mP_i + \mQ^l_i \, \mA = \mP_i + \sum_{j=1}^{p/c} \mQ^l_{ij}  \, \mA_j $.
However, the sum computed by $P(i,:)$ is only computes a partial sum. The results of these partial sums are then added with an all-reduce call on $P(i,:)$, yielding the correct $\mP_i$ matrix on each process in $P(i,j)$. If $q=p/c^2$, then the computation done on process $P(i,j)$ is $\mP_i = \mP_i + \mQ^l_i \, \mA = \mP_i + \sum_{k=jq}^{(j + 1)q} \mQ^l_{ik}  \, \mA_k$.
Thus, in each step of the summation, we must communicate $\mA_k$ to each process in $P(i,:)$. Our algorithm has $q$ steps, which broadcast successive $\mA_k$ block rows. 

Broadcasting the entire $\mA_k$ is a \textit{sparsity-oblivious} approach. This approach is simple and shows good scaling~\cite{spdmmm16,tripathy2020reducing}. However, $\mQ^l_{ik}$ is sparse, and sparsity-unaware SpGEMM algorithms unnecessarily communicate rows of $\mA_k$ that will not be read in the local $\mQ^l_{ik}\mA_k$ multiplication (i.e., whenever a column of $\mQ^l_{ik}$ has all zeros). For our 1.5D SpGEMM algorithm, we use a \textit{sparsity-aware} approach~\cite{ballard2013communication}. In this scheme, rather than broadcasting the entire block row $\mA_k$, we send to each process in $P(i,:)$ the specific rows needed for its local SpGEMM call.  Algorithm~\ref{alg:blockrow15dforward} shows pseudocode for our algorithm. 

\paragraph*{Communication Analysis}
\label{sec:commanalysis}
Both GraphSAGE and LADIES have the same communication costs in the first layer, so we consolidate both analyses for space. In addition, while we focus on the first layer, these analyses can be straightforwardly generalized to arbitary layers. 

In each iteration of the outer loop, there are $kb/(p/c)$ nonzero column ids gathered onto a process. The bandwidth cost of this step is $kb/c$, which is negligble compared to the cost of sending row data. 

In each iteration of the outer loop, there are $kb/(p/c)$ total rows sent to $P(:,j)$, each of which has $d$ nonzeros on average. The communication cost for sending row data $T_{rowdata} = \alpha(\log\frac{p}{c^2}) + \beta(\frac{kbd}{c})$
Finally, the all-reduce reduces matrices with $kbd/(p/c)$ nonzeros. This makes the communication cost for all-reduce $T_{allreduce} = \alpha\log c + \beta(\frac{ckbd}{p})$

In total, the communication cost for generating probability distributions in GraphSAGE and LADIES with our 1.5D algorithm is $T_{prob} = \alpha(\frac{p}{c^2} + \log c) + \beta(\frac{kbd}{c} + \frac{ckbd}{p})$.
\begin{algorithm}[t]
\begin{algorithmic}[1]
\STATE $s = p / c^2$  \COMMENT{number of stages}
\FOR{$q = 0$ to $s - 1$}
    \STATE $k = j \, s + q$
    \STATE $\vC \gets \textit{NnzCols}(\mQ^{l}_{ik})$
    \STATE $\hat{\vC} \gets \textit{Gather}(\vC, P(k,j))$
    \IF{$P(i,j) = P(k,j)$}
        \FOR{$l = 0$ to $p/c$}
            \STATE $\textit{ISend}(\mA[\hat{\vC}[l], :], P(l,j))$
        \ENDFOR
    \ENDIF
    \STATE $\textit{Recv}(\hat{\mA}, P(k,j))$
    \STATE $\hat{\mP} \gets \hat{\mP} + \textit{SpGEMM}(\mQ^{l}_{ik},\hat{\mA})$
\ENDFOR
\STATE $\mP^l \gets $ \textit{AllReduce}($\hat{\mP}$, +, $P(i, :)$)
%
\end{algorithmic}
\caption{Block 1.5D algorithm for generating probability distributions to sample from, which computes $\mP^l \gets \mQ^l\mA$ and normalizes each row depending on the sampling algorithm. $\mQ^{l}$ and $\mA$ are distributed on a $p / c \times c$ process grid.} \label{alg:blockrow15dforward}
\vspace{-0.1cm}
\end{algorithm}

We see from $T_{prob}$ that our 1.5D algorithm scales with the harmonic mean of $p/c$ and $c$.
\subsubsection{Sampling from Distributions}
Note that each distribution to sample is a separate row in $\mP$ and that $\mP$ is partitioned into block rows. Thus, sampling does not require communication.
\subsubsection{Row and Column Extraction}
For GraphSAGE, each process $P(i,j)$ only needs to manipulate it's local copy of $\mA_S$. No communication is necessary in this step, or extra steps for distribution. 

For LADIES, row extraction is SpGEMM between $\mQ^l_R\mA$. We use the same algorithm to run $\mQ^l_R\mA$ as $\mQ^l\mA$ outlined in Algorithm~\ref{alg:blockrow15dforward}. Here, $\mQ^l_R$ has dimensions $kb\times n$ with one nonzero per row. 

For LADIES column extraction, we run the SpGEMM $\mA_R\mQ^{l-1}_C$. Note that each submatrix $\mA_{Ri}$ is only multiplied with its respective column extraction matrix $\mQ^{l-1}_{Ci}$. Thus, we can interpret this large SpGEMM as a batch of smaller SpGEMMs, which are split across the process row followed by an all-reduce to avoid redundant work. In practice, a single process may also split its column extraction SpGEMM into smaller SpGEMM operations. This is because the $\mQ_C$ matrix has many empty rows, resulting in excessive memory costs to store the entire matrix in CSR form.
\revision{
\section{End-to-End Pipeline}
In this section, we describe our end-to-end training pipeline and how we implement each of three steps: 1) sampling, 2) feature fetching, and 3) propagation. These steps compose many other GNN training systems as well, and each system implements these steps differently.~\cite{kaler2023salient++,cai2023dsp,yang2022wholegraph,yang2022gnnlab}.}

\begin{figure}[t]
    \centering
    \includegraphics[scale=0.25]{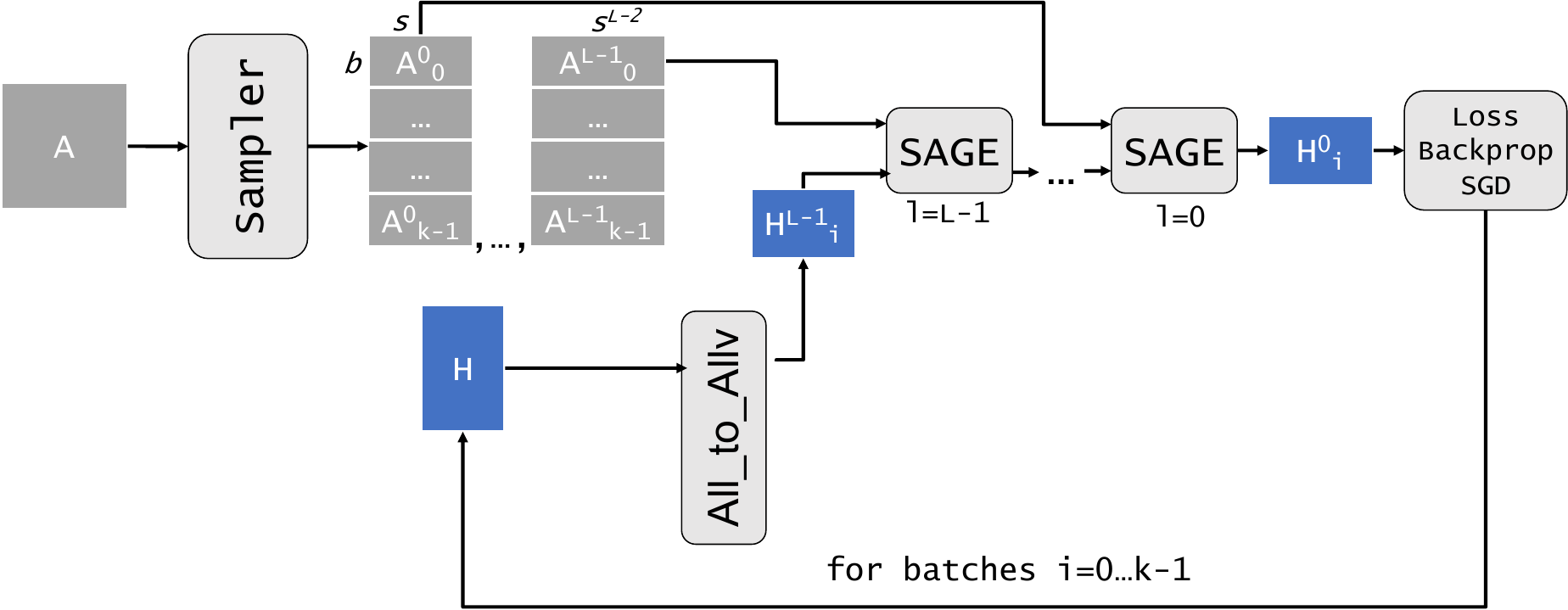}
    \caption{The overall architecture of our distributed pipeline from the perspective of one process. In this diagram, $\mA$ is either the entire adjacency matrix or a partition, depending on the algorithm, and $\mH$ is a partition of the input feature matrix. The first step in the pipeline is to run sampling on $k$ minibatches at once. Then, for each minibatch, we iteratively call \textit{all-to-allv} fetch the appropriate feature vectors, and run propagation on minibatch.}
    \label{fig:pipeline}
\vspace{-1.0cm}
\end{figure}

\revision{Figure~\ref{fig:pipeline} illustrates our pipeline from the perspective of a single GPU. The input feature matrix $\mH$ is partitioned with a 1.5D partitioning scheme. The input adjacency matrix $\mA$ is fully replicated on each GPU in our Graph Replicated algorithm, or partitioned with a 1.5D scheme in our Graph Partitioned algorithm. In addition, we treat our GPUs as $P(:)$ --- a 1.5D process grid. This is the same scheme used in Equation~\ref{eqn:15dpartitioning}, where both matrices are partitioned into block rows and each block row exists on the $c$ GPUs in its process row. 
Our pipeline runs bulk synchronously, where all GPUs participate in a single step simultaneously before advancing to the next step together. We discuss each step of our pipeline in the following sections.}

\revision{
\subsection{Sampling Step}
At the start of each epoch, each GPU begins by sampling $k/p$ total minibatches using either our Graph Replicated or Graph Partitioned algorithm. If $k$ is smaller than the total number of batches in our training set, we return later to sample remaining batches. The output of this step is each minibatch's sampled adjacency matrix for each layer .
}

\revision{\subsection{Feature Fetching and Propagation Steps}
Recall that neighborhood aggregation in forward propagation for a batch $i$ is an SpMM between a sampled adjacency matrix $\mA_i$ and a sampled feature matrix $\mH_i$. The sampled matrix $\mH_i$ contains the feature vectors (i.e. rows of $\mH$) of the vertices last frontier selected in the sampling step. }

\revision{After the sampling step, each GPU begins with the final layer's stacked matrix $\mA_S$ that contains $k/p$ sampled adjacency matrices to train with. Each process extracts a minibatch's sampled adjacency matrix $\mA_i$ from $\mA_S$ in a training step, and must collect the necessary rows of $\mH$ to form $\mH_i$ and begin training. To ensure each GPU has the necessary rows of $\mH$, all GPUs participate in an \textit{all-to-allv} call across process columns $P(:,j)$. Note that each process column contains the entirety of $\mH$, and a process $P(i, j)$ need only communicate with processes in its process column $P(:,j)$ to retrieve the rows of $\mH$ needed to begin forward propagation. With this scheme, our feature fetching time scales with the replication factor $c$. In our experiments, we increase $c$ as we increase the total number of processes $p$ since increasing the total number of processes also increases the total aggregate memory.}

\revision{Once each GPU has extracted a minibatch's adjacency matrix $\mA_i$ and fetched the rows needed to construct matrix $\mH_i$, we run forward and backward propagation on this minibatch. We repeat the $\textit{all-to-allv}$ step and propagation steps for all $k/p$ minibatches. If $k$ is less than the total number of minibatches in our training set, each GPU returns to the sampling step and repeats the pipeline until all batches in an epoch are trained.}
\section{Experimental Setup}
\subsection{Datasets}
We ran our experiments for both our Graph Replicated and Graph Partitioned algorithms on the datasets outlined in Table~\ref{tab:datasets}.  We borrow both our Protein dataset from CAGNET. Protein is originally from the HipMCL data repository~\cite{hipmcl}, and represents $1/8$th of it original vertices. This dataset are the largest studied in CAGNET and has randomly generated feature data for the purpose of measuring performance. In addition to Protein, we collect results on the Products and Papers datasets. Both are from the Open Graph Benchmark by Hu et. al., with Papers being the largest node-classification graph in the benchmark.~\cite{hu2020ogb}. We use a directed representation of Papers as opposed to an undirected version due to memory constraints with our Graph Replication algorithm. However, we achieve the correct accuracy as a serial execution of SAGE on directed Papers. \revision{Feature data for each dataset is stored with fp32, although lower precision features are feasible as well.}

In addition, we show scaling results on all three datasets for the architectures outlined in Table~\ref{tab:archicectures}. These hyperparameters are the most common choices for these datasets (e.g. in ~\cite{kaler2023salient++}). For LADIES, we restrict the model to one layer due to memory constraints. Our column extraction matrix has $kn$ rows and $n$ columns, which makes a CSR representation of this matrix memory-intensive. Since both cuSPARSE and nsparse both only support CSR-based SpGEMM, we must implement our column extraction SpGEMM with severall smaller CSR SpGEMMs that fit just below the memory cosntraints on our GPUs. This restriction it outlined in more detail in Section \ref{sec:ladiesperf}. We verify that the accuracy on each dataset matches existing works except for Protein, as the Protein dataset has randomly generated features for the purpose of measuring performance.
\begin{table}[t]
\centering
\caption{Datasets used in our experiments}
\label{tab:data}
\begin{tabular}{ |l|r|r|r|r|r|}
\hline
Name & Vertices & Edges & Batches & Features \\
\hline
Products & 2.4M & 126M & 196 & 100 \\
Protein & 8.7M & 1.3B & 1024 & 128 \\
Papers & 111M & 1.6B & 1172 & 128 \\
\hline
\end{tabular}
\label{tab:datasets}
\vspace{-0.2cm}
\end{table}
\begin{table}[t]
\centering
\caption{Architecture Parameters used in our experiments}
\begin{tabular}{ |l|r|r|r|r|r|}
\hline
GNN & Batch Size & Fanout & Hidden & Layers \\
\hline
SAGE & 1024 & (15,10,5) & 256 & 3 \\
LADIES & 512 & 512 & 256 & 1 \\
\hline
\end{tabular}
\label{tab:archicectures}
\vspace{-0.5cm}
\end{table}
\subsection{System Details}
We run all our experiments on the Perlmutter supercomputer at NERSC. Perlmutter GPU nodes have a  AMD EPYC 7763 (Milan) CPU and four NVIDIA A100 GPUs. Each pair of the GPUs has NVLINK 3.0 to communicate data at $100$GB/s unidirectional bandwidth. Each GPU has also $80$GB of HBM with $1552.2$GB/s memory bandwidth. Perlmutter nodes also have 4 HPE Slingshot 11 NICs, each with $25$GB/s injection bandwidth.
\subsection{Implementation Details}
We implement our framework in PyTorch 1.13~\cite{paszke2019pytorch} and PyTorch Geometric 2.2.0 with CUDA 11.5. We use NCCL 2.15 as our communications library, which has been widely used for GPU communication in deep learning problems~\cite{ncclRepo}. For our SpGEMM calls, we use the nsparse library~\cite{nsparse}. We report timings with the highest possible replication factor ($c$) and bulk minibatch count ($k$) without going out of memory for each GPU count. 
In the event where $k$ is less than the total number of minibatches as outlined in Table~\ref{tab:archicectures}, our pipeline repeats sampling the remaining minibatches in bulks of size $k$ to ensure each minibatch is trained on. In addition, we use Quiver as our baseline~\cite{quiver} for GraphSAGE. Quiver is a state-of-the-art GNN tool built on PyG that and is one of the only tools capable of handling large graphs with GPU sampling. We compare against Quiver GPU-only sampling, which fully replicates the graph on each device. For LADIES, to our knowledge, there does not exist a multi-node and multi-GPU distributed implementation. Thus, we only provide scaling numbers for LADIES in our framework.

\section{Results}

\subsection{Graph Replication Analysis}
\subsubsection{Quiver Comparison}
\begin{figure}[!t]
    \centering
    \includegraphics[scale=0.20]{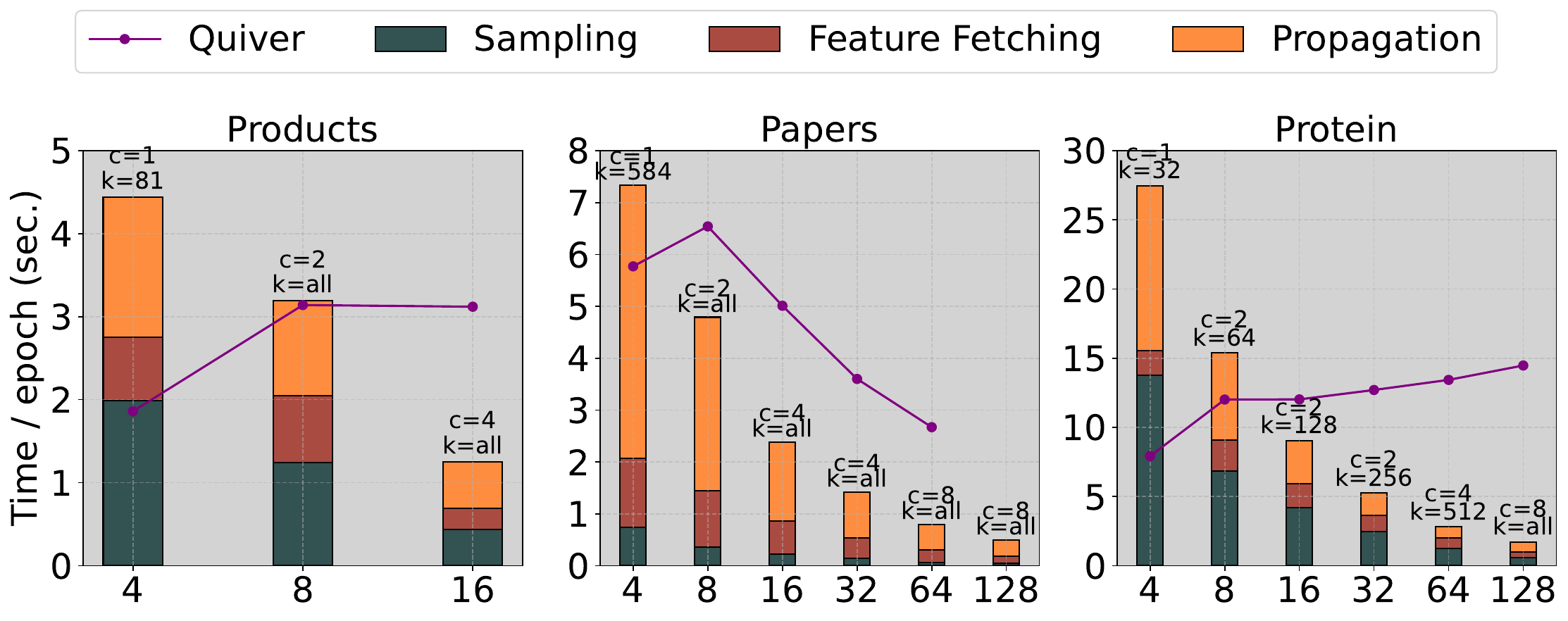} \caption{Performance results for our pipeline using the Graph Replication algorithm with GraphSAGE and compared with Quiver. For each GPU count, we break down the running time into 1) time spent sampling minibatches, 2) time spent in the feature fetching \textit{all-to-allv} call, and 3) time spent on forward and backward propagation.}
    \label{fig:graphrepbreakdown}
\end{figure}

Figure~\ref{fig:graphrepbreakdown} shows the total time taken by Quiver and our pipeline for each dataset, along with a performance breakdown for our pipeline. We show speedups over Quiver on large GPU counts on each dataset. On Products, we have a $2.5\times$ speedup over Quiver with $16$ GPUs, on Papers we have a $3.4\times$ speedup over Quiver on $64$ GPUs, and on Protein we have a $8.5\times$ speedup with $128$ GPUs. Quiver's preprocessing step ran out of memory on Papers with $128$ GPUs, so we do not include a Quiver datapoint there.

\begin{figure}[!t]
    \centering
    \includegraphics[scale=0.20]{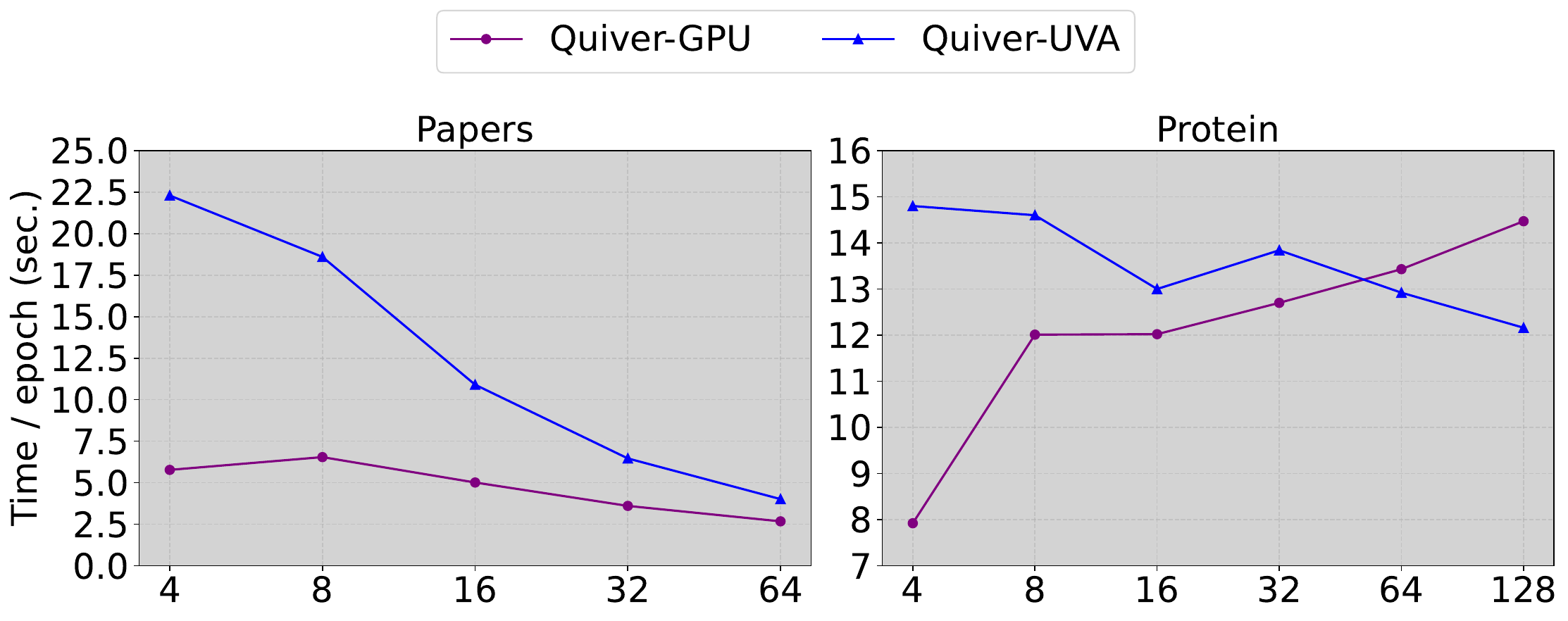} \caption{\revision{Performance comparison of Quiver with GPU sampling and Quiver training with Unified Virtual Address (UVA) sampling on Papers and Protein. UVA sampling uses both the CPU and multiple GPUs to run sampling.}}
    \label{fig:uvavgpu}
\end{figure}
\begin{figure}[!t]
    \centering
    \includegraphics[scale=0.20]{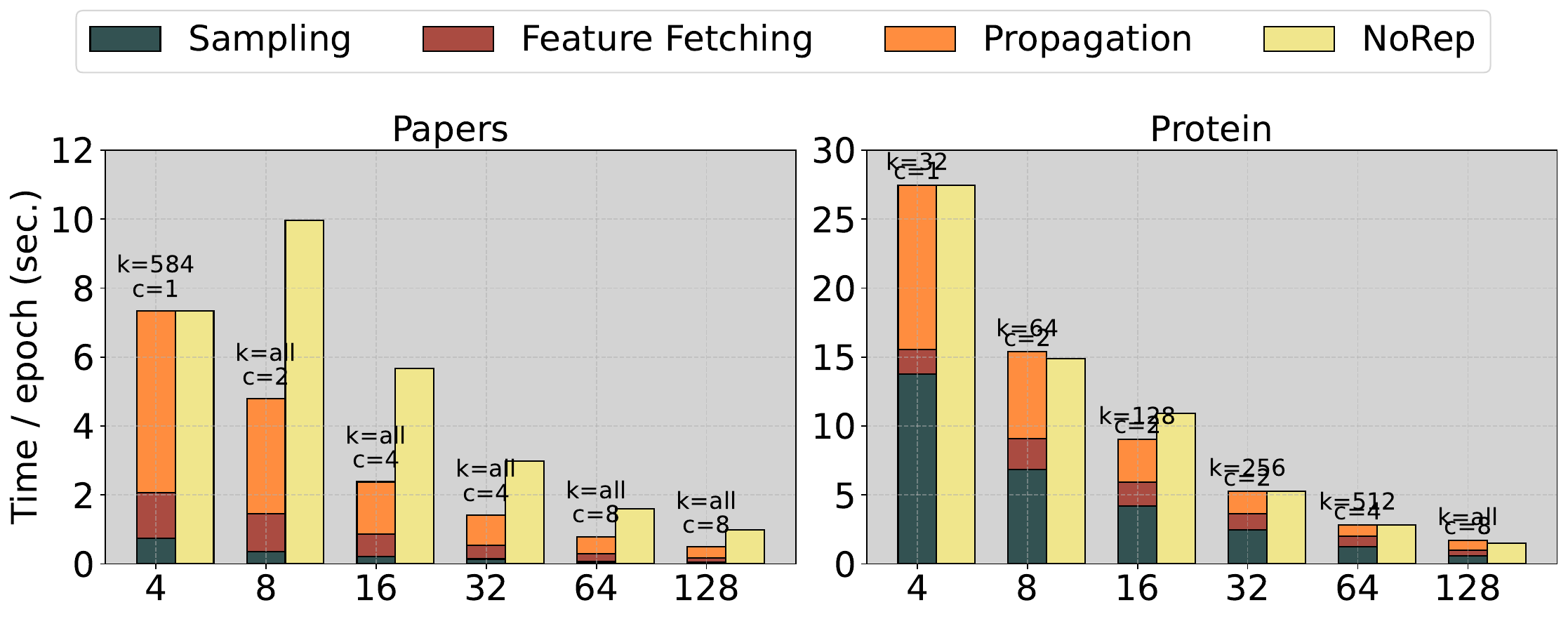} \caption{\revision{Performance results for our pipeline using the Graph Replication algorithm with GraphSAGE without replication on Papers and Protein.}}
    \label{fig:noreplication}
    \vspace{-0.2cm}
\end{figure}
Quiver experiences a slowdown on all datasets going from $4$ to $8$ GPUs. This is likely attributable to cross-node communication, as each node on Perlmutter has $4$ GPUs. 

Past $8$ GPUs, Quiver only scales on Papers, and does not scale at all on Products or Protein. This is likely because both Protein and Products are denser graphs, with an average degree of $241$ and $53$ compared to Papers ($29$). As a consequence, many features need to be communicated in the feature-fetching step, and Quiver does not effectively optimize this communication. This communication volume also increases as $p$ increases, causing Quiver to spend more time communicating and not scaling with $p$. Our pipeline is able to outperform to Quiver on large GPU counts since our feature fetching step scales with our replication factor $c$, along with our optimizations to sampling.

On low GPU counts, e.g. $4$ GPUs, we do not necessarily outperform Quiver particularly on Products and Protein. The advantages of our feature fetching approach come to fruition on larger GPU counts, as we do not have enough aggregate memory on lower GPU counts to replicate the feature matrix. In addition, our sampling optimizations are intended to amortize the cost of sampling over many minibatches. With only $4$ GPUs, we do not have enough memory to sample enough minibatches in bulk on Products and Protein to fully take advantage of our optimization. Note that in Figure~\ref{fig:graphrepbreakdown}, Products and Protein have a $k$ value smaller than their total minibatch count on $4$ GPUs. However, as we add more GPUs and aggregate memory, we are able to sample all minibatches in bulk for both Products and Protein. We are able to fully take advantage of our sampling operations at this point. For Papers, we are able to make sampling a small fraction of the total overall runtime on $4$ GPUs (only 10\% of the total time is spent on sampling). Papers has a high-vertex count and a low-density. Thus, Papers has many minibatches to sample from, and each minibatch takes less space on device than a minibatch from Products or Protein. For this reason, we are able to sample a large number of batches in bulk on Papers with few GPUs, yielding good performance by our sampling optimization even on only $4$ GPUs.

\revision{In addition, in Figure~\ref{fig:uvavgpu}  we include a comparison between Quiver's GPU and Unified Virtual Address (UVA) sampling, which stores the input graph in DRAM, and runs sampling on GPUs with a unified address space, and stores 80\% of features on DRAM with 20\% cached in GPU memory. For most GPU counts, the training time with GPU sampling outperforms that of UVA sampling. This comparison shows the benefits of GPU sampling over UVA or CPU sampling (the latter shows over a magnitude slowdown compared to UVA sampling). As the number of GPUs increases, the gap between UVA and GPU sampling shrinks, as sampling becomes a smaller fraction of the training time with additional GPUs.}
%
%

\subsubsection{Scaling Analysis}
In Figure~\ref{fig:graphrepbreakdown}, we see good scaling on all datasets. We have an 88\% parallel efficiency on Products, a 47\% parallel efficiency on Papers, and a 61\% parallel efficiency on Protein. Our sampling step scales nearly linearly with a $15.8\times$ speedup going from $4$ GPUs to $64$ GPUs on both Papers and Protein. This is because our sampling step in the Graph Replication algorithm involves no communication, and all minibatches in the $k$ to be sampled are partitioned across GPUs. Thus, the work per process decreases linearly, as evidenced by the linear scaling in sampling time. Our feature fetching time only decreases when we increase $c$. On Papers and Protein, our feature fetching time has a  $5.5\times$ speedup in feature fetching time when increasing $c$ by $8\times$ Papers, and a $4.5\times$ speedup when increasing $c$ by $8\times$ on Protein. Our propagation time scales linearly as the number of minibatches trained per GPU goes down linearly with GPU count.

\revision{In Figure~\ref{fig:noreplication}, we compare our pipeline's training times with replication to training times without replication. We can see the benefits of replication, with the performance degrading over $2\times$ without replication on Papers. This slowdown comes from increases in both the sampling phase and feature fetching phase. For Protein, we do not see significant benefits from replication. This is likely because the results in Figure~\ref{fig:graphrepbreakdown} never had a replication factor exceed $c>2$ due to memory constraints. In addition, for Protein, we were not able to sample all minibatches in bulk for most GPU counts due to memory constraints. Thus, a large portion of the runtime is likely sampling overheads that replication would not affect. Note that we use a naive replication scheme that simply partitions the feature matrix. Our pipeline could be improved by using sophisticated vertex caching schemes, such as those presented in SALIENT++~\cite{kaler2023salient++}}.

\subsubsection{Model Accuracies}
\revision{The optimizations we propose do not affect model accuracy. Our experiments use a $3$-layer SAGE model with PyG's implementation, although our methods support any model. On Products, with a batch size of $1024$, training fanout of $(15, 10, 5)$, and test fanout of $(20, 20, 20)$, our model reaches $77.8\%$ test accuracy after $20$ epochs of training on $4$ GPUs. This is within $1\%$ of the GraphSAGE result on the Open Graph Benchmark~\cite{hu2020ogb}. On Papers with the same parameters, our model reaches a test accuracy of $44.9\%$, which is consistent with a CPU execution of Papers when treated as a directed graph.}

\begin{figure}[t]
    \centering
    \includegraphics[scale=0.26]{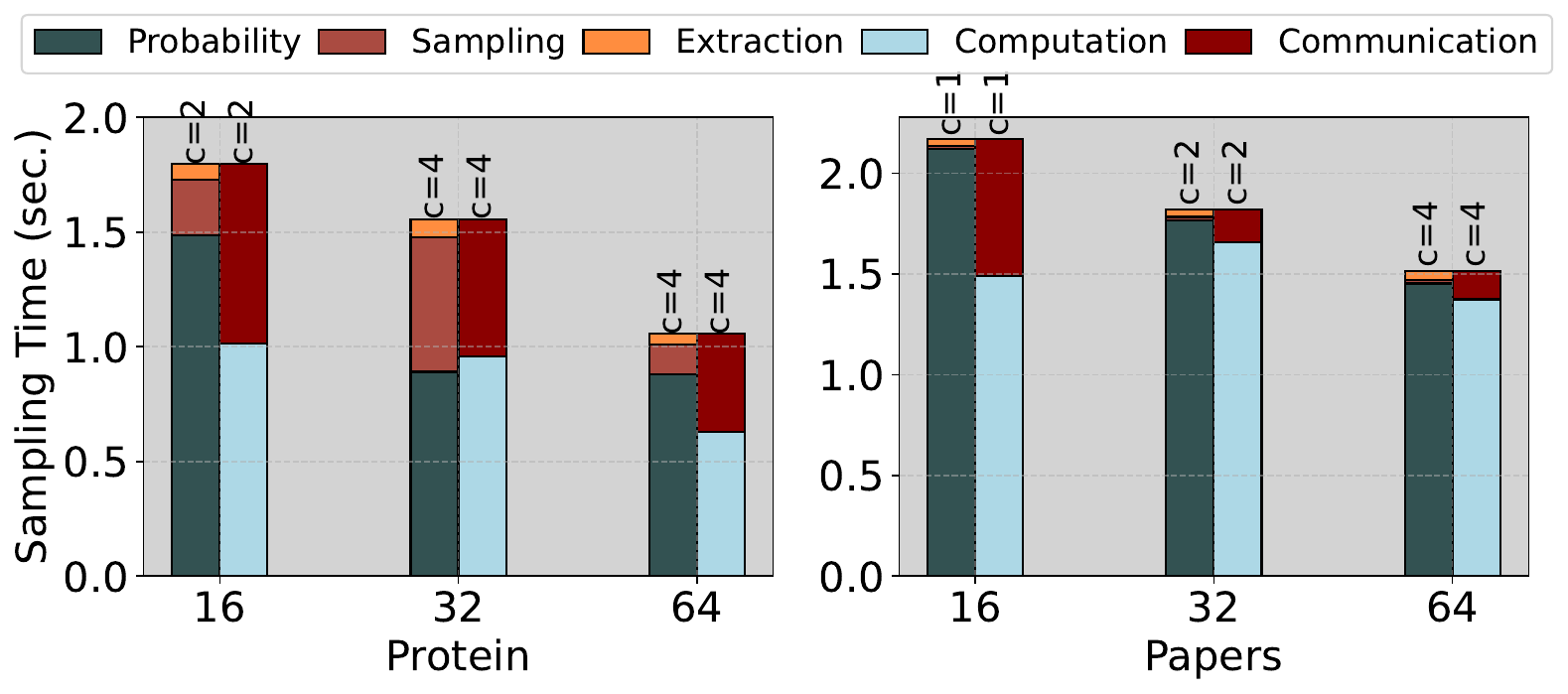}
    \includegraphics[scale=0.26]{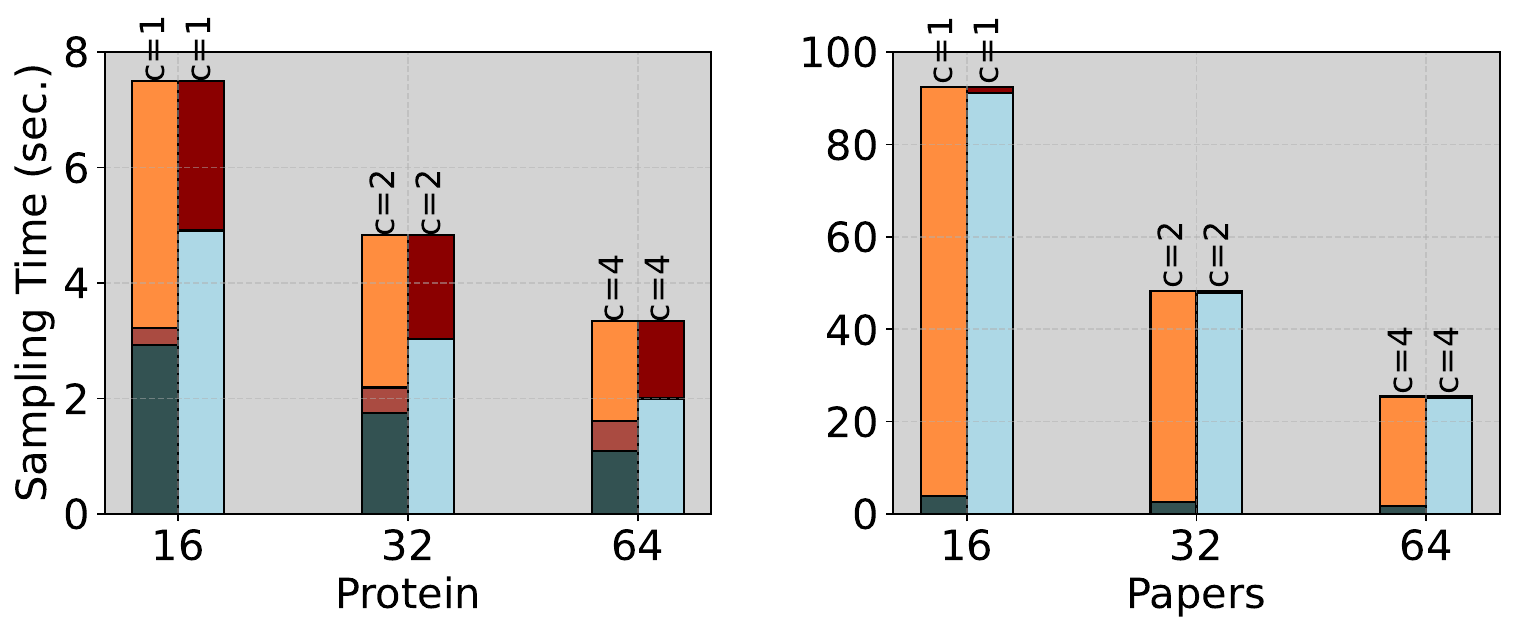}
    \caption{Scaling results for our Graph Partitioned algorithm. The first row of plots shows GraphSAGE timings, and the second row of plots shows LADIES timings. These plots only break down the time taken to sample minibatches, and breaks it down into the three steps outlined in Figures~\ref{fig:graphsage_figure},\ref{fig:ladies_figure}. We also break down the overall times into the time spent on communication and the time spent on computation. These times sample all minibatches in a single bulk sampling.}
    \label{fig:breakdown}
\end{figure} 
\subsection{Graph Partitioned Algorithm}
\subsubsection{GraphSAGE Performance Analysis}
Figure~\ref{fig:breakdown} shows the performance breakdown of our framework for our large datasets across process counts. We do not show results on Products as it is small enough to fit on device for most GPUs. 


Across all datasets, we see scaling up to $p=64$ GPUs. Protein sees $1.75\times$ speedup from $p=16$ to $p=64$, and Papers sees $1.43\times$ from $p=64$. This scaling behavior is consistent with the analysis in Section~\ref{sec:commanalysis}. Note that in Figure~\ref{fig:breakdown}, a majority of time is spent on probability computation, which is a sparsity-aware 1.5D SpGEMM. We discuss the scaling results by analyzing communication and computation for the 1.5D SpGEMM separately. 

Our analysis in Section~\ref{sec:commanalysis} shows that the communication time consists of $T_{rowdata}$ for sending row data and $T_{allreduce}$ for the final all-reduce call. The former scales with $c$, while the latter scales with $p/c$. Since $p > c$, significantly more time is spent communicating row data. Since sending row data is the bottleneck in communication, communication overall scales when $c$ increases. For a fixed $p$, increasing $c$ speeds up the SpGEMM. If $c$ cannot increase due to memory constraints, communication does not scale. This finding is consistent with prior work as well~\cite{buluc2008challenges}. Bulu\c{c} and Gilbert show that 1D SpGEMM algorithms are unscalable, where time increases with $p$. 
Computation, on the other hand, largely scales with $p$. Both sampling and extraction are computation steps, and these are embarrassingly parallel across $p$.
\subsubsection{LADIES Performance Analysis}
\label{sec:ladiesperf}
For our LADIES results in Figure~\ref{fig:breakdown}, we also see scaling across all datasets. Here, the time is dominated by the computation necessary for column extraction. \revision{In our implementation, we implement the column extraction step as a series of smaller SpGEMM calls as opposed to a single SpGEMM due to memory constraints. Our column extraction matrix is hypersparse and has $nk$ rows, making CSR storage memory intensive compared to COO or CSC. However, as cuSPARSE and nsparse support CSR-based SpGEMM, we split our column extraction matrix into smaller CSR matrices and smaller SpGEMM calls. For comparison, the reference CPU implementation for LADIES takes $43.9$ seconds to sample all minibatches for Papers and $3.12$ seconds for Protein. Our methods begin to exceed these times at $64$ GPUs.}
\section{Conclusion}
In this work, we introduce a matrix-based approach to sample minibatches in bulk. This approach outperforms exists distributed GNN libraries by (1) amortizing the cost of sampling minibatches, and (2) scaling communication by using communication-avoiding SpGEMM algorithms. In the future, we hope to express additional sampling algorithms in this framework. 
\section{Acknowledgements}
We thank our shepherd and reviewers for their constructive insight and feedback. This work is supported in part by the Advanced Scientific Computing Research (ASCR) Program of the Department of Energy Office of Science under contract
No. DE-AC02-05CH11231, and in part by the Exascale Computing Project
(17-SC-20-SC), a collaborative effort of the U.S. Department
of Energy Office of Science and the National Nuclear Security
Administration. This material is also based upon work supported the National Science Foundation 
under Award No. 1823034. 

This research used resources of the National Energy Research Scientific Computing Center at
the Lawrence Berkeley National Laboratory, which is supported by the Office of Science
of the U.S. Department of Energy under Contract No. DE-AC05-00OR22725.

\vspace{12pt}
\bibliography{gnnsampling}

\begin{thebibliography}{36}
\providecommand{\natexlab}[1]{#1}
\providecommand{\url}[1]{\texttt{#1}}
\expandafter\ifx\csname urlstyle\endcsname\relax
  \providecommand{\doi}[1]{doi: #1}\else
  \providecommand{\doi}{doi: \begingroup \urlstyle{rm}\Url}\fi

\bibitem[Azad et~al.(2018)Azad, Pavlopoulos, Ouzounis, Kyrpides, and
  Bulu{\c{c}}]{hipmcl}
Azad, A., Pavlopoulos, G.~A., Ouzounis, C.~A., Kyrpides, N.~C., and
  Bulu{\c{c}}, A.
\newblock {HipMCL: a high-performance parallel implementation of the Markov
  clustering algorithm for large-scale networks}.
\newblock \emph{Nucleic Acids Research}, 46\penalty0 (6):\penalty0 e33--e33, 01
  2018.
\newblock \doi{10.1093/nar/gkx1313}.
\newblock URL \url{https://doi.org/10.1093/nar/gkx1313}.

\bibitem[Ballard et~al.(2013)Ballard, Bulu{\c{c}}, Demmel, Grigori, Lipshitz,
  Schwartz, and Toledo]{ballard2013communication}
Ballard, G., Bulu{\c{c}}, A., Demmel, J., Grigori, L., Lipshitz, B., Schwartz,
  O., and Toledo, S.
\newblock Communication optimal parallel multiplication of sparse random
  matrices.
\newblock pp.\  222--231, 2013.

\bibitem[Bulu{\c{c}} \& Gilbert(2011)Bulu{\c{c}} and
  Gilbert]{bulucc2011combinatorial}
Bulu{\c{c}}, A. and Gilbert, J.~R.
\newblock The {C}ombinatorial {BLAS}: Design, implementation, and applications.
\newblock \emph{The International Journal of High Performance Computing
  Applications}, 25\penalty0 (4):\penalty0 496--509, 2011.

\bibitem[Bulu\c{c} \& Gilbert(2008)Bulu\c{c} and Gilbert]{buluc2008challenges}
Bulu\c{c}, A. and Gilbert, J.~R.
\newblock Challenges and advances in parallel sparse matrix-matrix
  multiplication.
\newblock In \emph{{The 37th International Conference on Parallel Processing
  (ICPP'08)}}, pp.\  503--510, Portland, Oregon, USA, September 2008.
\newblock \doi{10.1109/ICPP.2008.45}.
\newblock URL \url{http://eecs.berkeley.edu/~aydin/Buluc-ParallelMatMat.pdf}.

\bibitem[Cai et~al.(2023)Cai, Zhou, Yan, Zheng, Song, Zheng, Cheng, and
  Karypis]{cai2023dsp}
Cai, Z., Zhou, Q., Yan, X., Zheng, D., Song, X., Zheng, C., Cheng, J., and
  Karypis, G.
\newblock Dsp: Efficient gnn training with multiple gpus.
\newblock PPoPP '23, pp.\  392–404, 2023.

\bibitem[Cao et~al.(2023)Cao, Deng, Wu, Huang, Subbian, and
  Leskovec]{cao2023communicationfree}
Cao, K., Deng, R., Wu, S., Huang, E.~W., Subbian, K., and Leskovec, J.
\newblock Communication-free distributed gnn training with vertex cut, 2023.

\bibitem[Chen et~al.(2018)Chen, Ma, and Xiao]{chen2018fastgcn}
Chen, J., Ma, T., and Xiao, C.
\newblock Fastgcn: fast learning with graph convolutional networks via
  importance sampling.
\newblock \emph{arXiv preprint arXiv:1801.10247}, 2018.

\bibitem[Corporation(2023)]{ncclRepo}
Corporation, N.
\newblock {NCCL}: Optimized primitives for collective multi-gpu communication.
\newblock \url{https://github.com/NVIDIA/nccl}, 2023.

\bibitem[Fey \& Lenssen(2019)Fey and Lenssen]{pyg}
Fey, M. and Lenssen, J.~E.
\newblock Fast graph representation learning with {PyTorch Geometric}.
\newblock In \emph{ICLR Workshop on Representation Learning on Graphs and
  Manifolds}, 2019.

\bibitem[Gandhi \& Iyer(2021)Gandhi and Iyer]{gandhi2021p3}
Gandhi, S. and Iyer, A.~P.
\newblock P3: Distributed deep graph learning at scale.
\newblock In \emph{15th {USENIX} Symposium on Operating Systems Design and
  Implementation ({OSDI} 21)}, pp.\  551--568. {USENIX} Association, July 2021.
\newblock ISBN 978-1-939133-22-9.
\newblock URL
  \url{https://www.usenix.org/conference/osdi21/presentation/gandhi}.

\bibitem[Gholami et~al.(2018)Gholami, Azad, Jin, Keutzer, and
  Bulu\c{c}]{gholami2017integrated}
Gholami, A., Azad, A., Jin, P., Keutzer, K., and Bulu\c{c}, A.
\newblock Integrated model, batch, and domain parallelism in training neural
  networks.
\newblock In \emph{SPAA'18: 30th ACM Symposium on Parallelism in Algorithms and
  Architectures}, 2018.

\bibitem[Hamilton et~al.(2017)Hamilton, Ying, and
  Leskovec]{hamiltonInductive2017}
Hamilton, W., Ying, Z., and Leskovec, J.
\newblock Inductive representation learning on large graphs.
\newblock In Guyon, I., Luxburg, U.~V., Bengio, S., Wallach, H., Fergus, R.,
  Vishwanathan, S., and Garnett, R. (eds.), \emph{Advances in Neural
  Information Processing Systems 30}, pp.\  1024--1034. Curran Associates,
  Inc., 2017.
\newblock URL
  \url{http://papers.nips.cc/paper/6703-inductive-representation-learning-on-large-graphs.pdf}.

\bibitem[Hu et~al.(2020)Hu, Fey, Zitnik, Dong, Ren, Liu, Catasta, and
  Leskovec]{hu2020ogb}
Hu, W., Fey, M., Zitnik, M., Dong, Y., Ren, H., Liu, B., Catasta, M., and
  Leskovec, J.
\newblock Open graph benchmark: Datasets for machine learning on graphs.
\newblock \emph{arXiv preprint arXiv:2005.00687}, 2020.

\bibitem[H{\"u}bschle-Schneider \& Sanders(2022)H{\"u}bschle-Schneider and
  Sanders]{schneider2022parallel}
H{\"u}bschle-Schneider, L. and Sanders, P.
\newblock Parallel weighted random sampling.
\newblock volume~48, pp.\  1--40. ACM, 2022.

\bibitem[Jangda et~al.(2021)Jangda, Polisetty, Guha, and
  Serafini]{jangda2021accelerating}
Jangda, A., Polisetty, S., Guha, A., and Serafini, M.
\newblock Accelerating graph sampling for graph machine learning using gpus.
\newblock In \emph{EuroSys '21: Proceedings of the Sixteenth European
  Conference on Computer Systems}, pp.\  311--326. ACM, 2021.

\bibitem[Jia et~al.(2020)Jia, Lin, Gao, Zaharia, and Aiken]{mlsys2020_83}
Jia, Z., Lin, S., Gao, M., Zaharia, M., and Aiken, A.
\newblock Improving the accuracy, scalability, and performance of graph neural
  networks with {ROC}.
\newblock In \emph{Proceedings of Machine Learning and Systems (MLSys)}, pp.\
  187--198. 2020.

\bibitem[Jiao et~al.(2023)Jiao, Li, Wu, Hu, Li, Bian, Dai, Luo, Hu, Huang,
  Feng, Yang, Feng, Xiong, Yu, Li, He, Ma, and Liu]{pglbox}
Jiao, X., Li, W., Wu, X., Hu, W., Li, M., Bian, J., Dai, S., Luo, X., Hu, M.,
  Huang, Z., Feng, D., Yang, J., Feng, S., Xiong, H., Yu, D., Li, S., He, J.,
  Ma, Y., and Liu, L.
\newblock Pglbox: Multi-gpu graph learning framework for web-scale
  recommendation.
\newblock In \emph{Proceedings of the 29th ACM SIGKDD Conference on Knowledge
  Discovery and Data Mining}, KDD '23, pp.\  4262–4272, 2023.

\bibitem[Kaler et~al.(2023)Kaler, Iliopoulos, Murzynowski, Schardl, Leiserson,
  and Chen]{kaler2023salient++}
Kaler, T., Iliopoulos, A., Murzynowski, P., Schardl, T., Leiserson, C.~E., and
  Chen, J.
\newblock Communication-efficient graph neural networks with probabilistic
  neighborhood expansion analysis and caching.
\newblock In \emph{Proceedings of Machine Learning and Systems}, 2023.

\bibitem[Koanantakool et~al.(2016)Koanantakool, Azad, Bulu\c{c}, Morozov, Oh,
  Oliker, and Yelick]{spdmmm16}
Koanantakool, P., Azad, A., Bulu\c{c}, A., Morozov, D., Oh, S.-Y., Oliker, L.,
  and Yelick, K.
\newblock Communication-avoiding parallel sparse-dense matrix-matrix
  multiplication.
\newblock In \emph{Proceedings of the IPDPS}, 2016.

\bibitem[Ma et~al.(2019)Ma, Yang, Miao, Xue, Wu, Zhou, and Dai]{neugraph}
Ma, L., Yang, Z., Miao, Y., Xue, J., Wu, M., Zhou, L., and Dai, Y.
\newblock {NeuGraph}: Parallel deep neural network computation on large graphs.
\newblock In \emph{{USENIX} Annual Technical Conference ({USENIX} {ATC} 19)},
  pp.\  443--458, Renton, WA, 2019. {USENIX} Association.
\newblock ISBN 978-1-939133-03-8.

\bibitem[Nagasaka et~al.()Nagasaka, Nukada, and Matsuoka]{nsparse}
Nagasaka, Y., Nukada, A., and Matsuoka, S.
\newblock Adaptive multi-level blocking optimization for sparse matrix vector
  multiplication on gpu.
\newblock \emph{Procedia Comput. Sci.}, 80\penalty0 (C):\penalty0 131–142.
\newblock ISSN 1877-0509.
\newblock \doi{10.1016/j.procs.2016.05.304}.
\newblock URL \url{https://doi.org/10.1016/j.procs.2016.05.304}.

\bibitem[Olver \& Townsend(2013)Olver and Townsend]{olver2013fast}
Olver, S. and Townsend, A.
\newblock Fast inverse transform sampling in one and two dimensions.
\newblock \emph{arXiv preprint arXiv:1307.1223}, 2013.

\bibitem[Pandey et~al.(2020)Pandey, Li, Hoisie, Li, and Liu]{pandey2020csaw}
Pandey, S., Li, L., Hoisie, A., Li, X., and Liu, H.
\newblock C-saw: A framework for graph sampling and random walk on gpus.
\newblock In \emph{SC20: International Conference for High Performance
  Computing, Networking, Storage and Analysis}, pp.\  1--14. IEEE, 2020.

\bibitem[Paszke et~al.(2019)Paszke, Gross, Massa, Lerer, Bradbury, Chanan,
  Killeen, Lin, Gimelshein, Antiga, et~al.]{paszke2019pytorch}
Paszke, A., Gross, S., Massa, F., Lerer, A., Bradbury, J., Chanan, G., Killeen,
  T., Lin, Z., Gimelshein, N., Antiga, L., et~al.
\newblock {PyTorch}: An imperative style, high-performance deep learning
  library.
\newblock In \emph{Advances in Neural Information Processing Systems}, pp.\
  8024--8035, 2019.

\bibitem[Team(2023)]{quiver}
Team, Q.
\newblock Torch-quiver: Pytorch library for fast and easy distributed graph
  learning.
\newblock \url{https://github.com/quiver-team/torch-quiver}, 2023.

\bibitem[Tripathy et~al.(2020)Tripathy, Yelick, and
  Bulu{\c{c}}]{tripathy2020reducing}
Tripathy, A., Yelick, K., and Bulu{\c{c}}, A.
\newblock Reducing communication in graph neural network training.
\newblock In \emph{SC20: International Conference for High Performance
  Computing, Networking, Storage and Analysis}, pp.\  1--14. IEEE, 2020.

\bibitem[Wan et~al.(2022{\natexlab{a}})Wan, Li, Li, Kim, and Lin]{wan2022bns}
Wan, C., Li, Y., Li, A., Kim, N.~S., and Lin, Y.
\newblock {BNS-GCN}: Efficient full-graph training of graph convolutional
  networks with partition-parallelism and random boundary node sampling.
\newblock \emph{Proceedings of Machine Learning and Systems}, 4:\penalty0
  673--693, 2022{\natexlab{a}}.

\bibitem[Wan et~al.(2022{\natexlab{b}})Wan, Li, Wolfe, Kyrillidis, Kim, and
  Lin]{wan2022pipegcn}
Wan, C., Li, Y., Wolfe, C.~R., Kyrillidis, A., Kim, N.~S., and Lin, Y.
\newblock Pipe{GCN}: Efficient full-graph training of graph convolutional
  networks with pipelined feature communication.
\newblock In \emph{International Conference on Learning Representations},
  2022{\natexlab{b}}.
\newblock URL \url{https://openreview.net/forum?id=kSwqMH0zn1F}.

\bibitem[Wu et~al.(2020)Wu, Pan, Chen, Long, Zhang, and
  Philip]{wu2020comprehensive}
Wu, Z., Pan, S., Chen, F., Long, G., Zhang, C., and Philip, S.~Y.
\newblock A comprehensive survey on graph neural networks.
\newblock \emph{IEEE transactions on neural networks and learning systems},
  32\penalty0 (1):\penalty0 4--24, 2020.

\bibitem[Yang et~al.(2022{\natexlab{a}})Yang, Bulu{\c{c}}, and
  Owens]{yang2022graphblast}
Yang, C., Bulu{\c{c}}, A., and Owens, J.
\newblock Graphblast: A high-performance linear algebra-based graph framework
  on the gpu.
\newblock \emph{ACM Transactions on Mathematical Software}, 48\penalty0
  (1):\penalty0 1--51, 2022{\natexlab{a}}.

\bibitem[Yang et~al.(2022{\natexlab{b}})Yang, Liu, Qi, and
  Lai]{yang2022wholegraph}
Yang, D., Liu, J., Qi, J., and Lai, J.
\newblock Wholegraph: A fast graph neural network training framework with
  multi-gpu distributed shared memory architecture.
\newblock In \emph{Proceedings of the International Conference on High
  Performance Computing, Networking, Storage and Analysis}, SC '22,
  2022{\natexlab{b}}.

\bibitem[Yang et~al.(2022{\natexlab{c}})Yang, Tang, Song, Wang, Yin, Chen, Yu,
  and Zhou]{yang2022gnnlab}
Yang, J., Tang, D., Song, X., Wang, L., Yin, Q., Chen, R., Yu, W., and Zhou, J.
\newblock Gnnlab: A factored system for sample-based gnn training over gpus.
\newblock In \emph{Proceedings of the Seventeenth European Conference on
  Computer Systems}, EuroSys '22, pp.\  417–434, 2022{\natexlab{c}}.

\bibitem[Yang et~al.(2019)Yang, Zhang, Chen, Ma, Bai, and Jiang]{knightking}
Yang, K., Zhang, M., Chen, K., Ma, X., Bai, Y., and Jiang, Y.
\newblock Knightking: A fast distributed graph random walk engine.
\newblock 2019.

\bibitem[Zheng et~al.(2020)Zheng, Ma, Wang, Zhou, Su, Song, Gan, Zhang, and
  Karypis]{zheng2020distdgl}
Zheng, D., Ma, C., Wang, M., Zhou, J., Su, Q., Song, X., Gan, Q., Zhang, Z.,
  and Karypis, G.
\newblock Distdgl: distributed graph neural network training for billion-scale
  graphs.
\newblock In \emph{2020 IEEE/ACM 10th Workshop on Irregular Applications:
  Architectures and Algorithms (IA3)}, pp.\  36--44. IEEE, 2020.

\bibitem[Zhu et~al.(2019)Zhu, Zhao, Yang, Lin, Zhou, Ai, Li, and
  Zhou]{zhu2019aligraph}
Zhu, R., Zhao, K., Yang, H., Lin, W., Zhou, C., Ai, B., Li, Y., and Zhou, J.
\newblock {AliGraph}: a comprehensive graph neural network platform.
\newblock \emph{Proceedings of the VLDB Endowment}, 12\penalty0 (12):\penalty0
  2094--2105, 2019.

\bibitem[Zou et~al.(2019)Zou, Hu, Wang, Jiang, Sun, and Gu]{ladies}
Zou, D., Hu, Z., Wang, Y., Jiang, S., Sun, Y., and Gu, Q.
\newblock Layer-dependent importance sampling for training deep and large graph
  convolutional networks.
\newblock In \emph{Proceedings of Neural Information Processing Systems
  (NeurIPS)}, 2019.

\end{thebibliography}
\bibliographystyle{mlsys2024}

\end{document}